# Enhancing Wireless Device Identification through RF Fingerprinting: Leveraging Transient Energy Spectrum Analysis

**Nisar Ahmed[1], Gulshan Saleem[2], Hafiz Muhammad Shahzad Asif[1], Muhammad Usman Younus[3], Kalsoom Safdar[4]**

[1]Department of Computer Science, University of Engineering and Technology Lahore, New Campus, Punjab, Pakistan.
[2]Faculty of Information Technology & Computer Science, University of Central Punjab, Lahore, Punjab, Pakistan.
[3]Department of Computer Science & Information Technology, Baba Guru Nanak University, Nankana Sahib, Punjab, Pakistan.
[4]Department of Computer Science & Information Technology, University of Jhang, Jhang, Punjab, Pakistan.
For correspondence: nisarahmedrana@yahoo.com

**Abstract**: In recent years, the rapid growth of the Internet of Things technologies and the widespread adoption of 5G wireless networks have led to an exponential increase in the number of radiation devices operating in complex electromagnetic environments. A key challenge in managing and securing these devices is accurate identification and classification. To address this challenge, specific emitter identification techniques have emerged as a promising solution that aims to provide reliable and efficient means of identifying individual radiation devices in a unified and standardized manner. This research proposes an approach that leverages transient energy spectrum analysis using the General Linear Chirplet Transform to extract features from RF devices. A dataset comprising nine RF devices is utilized, with each sample containing 900 attributes and a total of 1080 equally distributed samples across the devices. These features are then used in a classification modeling framework. To overcome the limitations of conventional machine learning methods, we introduce a hybrid deep learning model called the CNN-Bi-GRU for learning the identification of RF devices based on their transient characteristics. The proposed approach provided a 10-fold cross-validation performance with a precision of 99.33%, recall of 99.53%, F1-score of 99.43%, and classification accuracy of 99.17%. The results demonstrate the promising classification performance of the CNN-Bi-GRU approach, indicating its suitability for accurately identifying RF devices based on their transient characteristics and its potential for enhancing device identification and classification in complex wireless environments.

## 1 Introduction

The rapid advancements of the Internet of Things (IoT) [1] and 5G [2] wireless technologies have led to a proliferation of interconnected wireless devices, resulting in various regulatory and security challenges. To address these issues and enhance IoT security, specific emitter identification (SEI) technology has been proposed for identity authentication and device supervision. Accurate identification of radiation source devices of the same type has become a critical problem in ensuring communication security and privacy.

SEI involves distinguishing individual transmitters by comparing radio frequency (RF) fingerprints present in received signals [3]. These RF fingerprints represent unique hardware-specific characteristics [4] inherent to the analog components of the transmitter and are independent of the transmitted information. SEI plays an increasingly significant role, particularly in cognitive radio and ad hoc networks. The difficulty of tampering with or forging RFFs makes them highly valuable for physical layer security, equipment identification, and certification [5, 6].

In a typical SEI system, as depicted in Figures 1 & 2, the process involves extracting RF fingerprints and performing device identification. SEI techniques can be broadly categorized based on the type of signals they target: transient signals and steady-state signals. Transient signals, characterized by an increasing signal power, provide a rich source of transmitter hardware-specific features, contributing to a large number of RFFs for SEI. The problem of identifying a wireless device among a group of devices [7] that use the same wireless standard, such as WiFi, is especially interesting. To tackle this challenge effectively, any proposed technique should possess desirable characteristics, including high identification accuracy and robustness to environmental effects in the wireless propagation medium, such as the presence of background RF noise or fading.

Common approaches to tackle this problem involve techniques such as Variational Mode Decomposition and Spectral Features (VMD-SF) [8], Hilbert transforms for RF fingerprint extraction [9], and the utilization of GLCT [10] for extracting transient features [11]. These features are then used to train classification models for wireless device identification. The most common classification choices are support vector machines, random forests, and artificial neural networks. Deep neural networks have recently received prominence [12-14] and [15-18] and have explored the use of CNN for wireless device identification based on RF fingerprinting.

Our research contribution addresses the challenge of wireless device identification using a dataset provided by [19], which consists of 1080 samples from nine different wireless devices. Each sample contains 900 features extracted using the GLCT from transient analysis of a signal. To tackle this problem, we have introduced a novel hybrid architecture combining Convolutional Neural Networks (CNN) with Bidirectional Gated Recurrent Units (Bi-GRU).

In contrast to existing solutions, our hybrid CNN-Bi-GRU architecture offers distinct advantages for modeling wireless device identification. We have also explored several other deep learning approaches, including CNN, Bi-LSTM, Bi-GRU, and CNN-Bi-GRU, to evaluate their suitability for this task. Through comprehensive experimentation, we have found that the proposed CNN-Bi-GRU architecture consistently outperforms all the alternative models, demonstrating its superior effectiveness in accurately identifying wireless devices.

Our proposed approach aims to enhance identification accuracy while also improving the robustness of the system against background noise which is evaluated by performing identification at various SNR levels. By effectively modeling the complex temporal and spatial characteristics of RF fingerprints, the CNN-Bi-GRU architecture provides a more reliable and robust solution for wireless device identification. This advancement in identification accuracy and robustness contributes to strengthening communication security and privacy, particularly in the context of IoT and 5G wireless environments.

## 2  Literature Review

Device identification and authentication play crucial roles in the management of wireless networks. However, the rapid proliferation of wireless devices in our environment has made these tasks increasingly challenging due to the growing number of emitters and potential attack surfaces [20-22]. Furthermore, the terms "identification" and "authentication" are often used interchangeably, leading to confusion and misunderstanding. In the broader context, identification can be viewed as a subtask within the authentication and authorization process. It involves uniquely identifying a device based on its unique ID. While device IDs are commonly used for authentication using secret keys, they can still pose vulnerabilities that may be exploited by unauthorized users to gain network access. RF fingerprinting, which captures hardware-specific and often unintentional characteristics at the analog component level, offers a promising approach to enhance identification and authentication. The inherent difficulty in mimicking RF fingerprints makes them resilient against impersonation. Therefore, incorporating RF fingerprinting into a more robust identification and authentication framework can strengthen the security of wireless networks.

To leverage the potential of RF fingerprints for wireless device identification, researchers have explored various aspects of transmitting equipment [23]. Previous studies have investigated a range of factors, such as statistical moments of amplitude, phase, and frequency profiles [24], transient behaviors during device turn-on [25, 26], carrier frequency offset [27], power amplifier non-linearity, and I/Q offset. These investigations have been motivated by the ability of these characteristics to provide unique traits that can aid in device identification [25, 28]. To enhance the accuracy of device identification, machine learning techniques [29], including support vector machines [24], decision trees [30], k-nearest neighbors, and artificial neural networks [31], have been widely adopted. While numerous RF fingerprints can be identified, tracking them back to their underlying features can pose challenges due to the complex nonlinear interactions among multiple underlying processes [32]. Consequently, attributing specific fingerprints to their corresponding features may prove to be a demanding task.

## 2.1 RF Fingerprinting Approaches

In the field of RF fingerprinting for wireless device identification, researchers have explored various approaches to extract unique features to enhance identification accuracy [5, 33, 34]. One approach focuses on the transient component of the signal that occurs during power-on, utilizing energy criterion and transient duration-based techniques [35]. Statistical parameters are down-selected in the signal preamble area to minimize computational costs while maintaining accuracy [35]. Another feature extraction approach involves representing time series data in a two-dimensional format [18].

Physical Unclonable Features (PUFs) are utilized to distinguish between different transmitters without compromising signal quality [36]. PUFs can be categorized as error patterns and transient patterns [36]. Error patterns exploit the statistical features of received symbol noise to uniquely characterize wireless devices [37]. Phase inaccuracy in transmitters' Phase Locked Loop (PLL) has been shown to have beneficial effects, where error vectors derived from theoretical models and observed signals can generate unique fingerprints for identification [37]. The authors of [18] capture time-varying modulation errors using a Differential Constellation Trace Image (DCTF) to identify various Zigbee devices.

Transient patterns, which indicate the start and finish of wireless packet transmissions, are resistant to manipulation by adversaries and are used to identify the source of emissions [25, 38]. Researchers have shown that the transient energy spectrum on the turn-on amplitude envelopes of transmitters, particularly in the frequency domain, provides informative characteristics [25]. While feature-based techniques require meticulous extraction of high-order statistics or features for each specific scenario, there is a need for less labor-intensive and more flexible approaches [35].

Multifractal analysis has been proposed to assess irregularity levels in transmitted signals and detect changes within this categorization [39]. Permutation entropy has been used to differentiate between noise and signals based on their complexity gaps [39]. The variance of phase characteristic differences has been employed for transient detection in wireless transceivers [39]. Bayesian methods, such as step data models and linear ramp data models, have been applied to identify transients by computing the posterior probability distribution of change points [40]. However, the computational cost associated with these Bayesian methods limits their applicability [40].

The energy criterion technique has been proposed for transient detection, classifying received signals based on variations in energy quantities [35]. Evaluations using Wi-Fi signals captured from smartphones and devices from different manufacturers have demonstrated the effectiveness of this technique [35, 41]. Various transient detection strategies have been extensively assessed using experimental data collected from a wide range of wireless transmitters [35, 41].

## 2.2 Transient-Based RF Fingerprinting

Transient signal properties such as amplitude, phase, and frequency have been investigated recently for their potential use in device identification. In one study, signal features such as normalized amplitude variance, number of peaks in the carrier signal, and transient energy spectrum were transformed into RF fingerprints [25]. Other investigations focused on converting transient signals to the wavelet domain and extracting RF fingerprinting features using wavelet analysis and multi-resolution analysis [3].

Recent advancements in feature extraction methods have provided more sophisticated approaches. For instance, one study retrieved RF fingerprints by performing a least-squares fit on the transient envelope and extracting the fitting coefficients [42]. In 2019, another approach employed a multimodal sparse representation technique to recover RF fingerprints [43]. Hilbert-Huang transform-based techniques have also gained attention in the academic community. A research group led by A. Kara utilized the Hilbert-Huang transform to analyze transient signals, examined the resulting time-frequency energy distribution, and extracted statistical properties as RF fingerprints [44]. To enhance the robustness of RF fingerprints to noise, the same research team utilized variational mode decomposition to reconstruct the transient signal [33]. Additionally, Yuan expanded upon the Hilbert-Huang transform approach by creating 13-dimensional features that capture both broad and subtle transient properties [45].

### *2.3 Emerging Trends in Wireless Device Identification*

Deep Neural Networks (DNNs) have gained popularity as a substitute for Blackbox methods in various domains [46-49], including device-specific identification [15, 16]. DNN-enabled wireless device identification systems often employ convolutional layers to uncover latent properties. These layers utilize filters to automatically gather potentially valuable information, eliminating the need for manual feature selection. Recent years have witnessed an upsurge in research on deep learning approaches for device identification, including methods like CNN [11, 17, 18, 50], Probabilistic Neural Networks (PNN) [6, 11, 17, 18, 25-27, 50], Autoencoders [15, 16], and Long Short-Term Memory (LSTM) [16]. These techniques offer the advantage of not requiring human feature engineering, eliminating the need for prior knowledge about fingerprint characteristics.

While the literature mainly focuses on digital communication technologies such as Wi-Fi [16, 25], Bluetooth [26], and Zigbee [15, 18], the available RF fingerprints depend on the technologies and system architectures used. Different technologies employ diverse modulation methods, packet formats, and standards, adding complexity to the system when adapting RF fingerprinting techniques for use with other systems.

In some cases, instead of manually selecting features, time series data is directly provided to CNN, allowing the network to learn the optimal feature set for the machine learning task. CNNs have been employed for classifying multiple devices under different operating conditions [35], verifying claimed device identification against a small pool of known devices, and evaluating intentional IQ imbalances created by transmitters. To enhance RF fingerprinting methods for real-world applications, researchers have explored approaches such as simplifying neural network operations for deployment on edge devices with limited resources [25]. The use of Siamese Networks has also been proposed to avoid constant retraining of RFF models when adding additional nodes to an existing network, thereby improving model scalability.

The literature addresses the impact of deployment unpredictability on research findings. For example, [3] delves into the variability resulting from factors such as time, location, and receiver configuration, while [43] explores the impact of the RF environment. In many cases, reducing or eliminating environmental variables is necessary to improve classification accuracy. Techniques such as data augmentation [44], pre-transmit filtering [33], intentional IQ mismatch, and channel equalization are viable strategies to achieve this goal.

Innovative methods combining CNNs with autoencoders have been proposed for signal denoising and emitter identification, even in low Signal-to-Noise Ratio (SNR) scenarios [51]. Comparable studies have achieved similar outcomes using stacked denoising autoencoders [52]. Deep neural networks have demonstrated effectiveness in processing unprocessed inputs, as highlighted by [53], where a Deep CNN outperforms traditional statistical learning methods without the need for feature engineering. Researchers in [54] utilized raw IQ scores from the CorteXlab dataset to train neural networks, obtaining consistent results with other studies. Deep neural networks enhance the quality of fingerprints while reducing the requirement for domain knowledge compared to feature-based techniques. DNNs are evolving into a practical component for non-cryptographic wireless device identification. However, anomaly detection remains challenging for DNNs, as it necessitates high performance on both learned and unknown data. These systems need to excel in identifying novel or unexpected data while performing accurate machine learning in practical settings.

## 2.4 Research Gap

Despite a wealth of research on fingerprinting and classification techniques, static classification methods are used in many current approaches for RF fingerprint-based wireless device identification. The transient nature of RF fingerprints, which change over time due to a variety of factors like temperature changes, hardware limitations, and environmental effects, is difficult for these methods to capture. Efficient modeling through conventional static techniques is severely hindered by this intrinsic dynamism.

Investigating the possibilities of deep learning architectures with temporal learning mechanisms is becoming more and more important in order to get around this limitation. LSTM and GRU networks are two types of Recurrent Neural Networks (RNNs) that have shown impressive abilities to capture temporal dependencies in sequential data.

However, because CNNs are better at learning complex feature maps from transient data, most existing research focuses on using CNNs for feature extraction from RF signals. CNNs are good at capturing spatial features, but they can perform even better when combined with temporal learning networks such as GRU or LSTM. By fusing the temporal learning capabilities of LSTM or GRU networks with CNN's proficiency in spatial feature extraction, this integration makes the most of both networks' strengths.

This synergy improves classification performance for wireless device identification by enabling a more thorough understanding of the dynamic properties of RF fingerprints. The efficiency of this method is further increased by pre-processing the RF signals using the GLCT. Compared to other Time Frequency Analysis (TFA) techniques like Wavelet transform, Short-Time Fourier Transform (STFT), Empirical Mode Decomposition, Hilbert-Huang Transform or Wigner-Ville Distribution, GLCT provides a richer and more reliable feature representation by effectively extracting informative features from non-stationary signals like RF emissions. The CNN-Bi-GRU model can learn more intricate and discriminative patterns due to this enhanced feature representation, which improves accuracy and robustness in wireless device identification tasks—especially when the signal-to-noise ratio is low.

We hypothesize that combining GLCT with a CNN-Bi-GRU architecture will result in state-of-the-art RF device detection performance by exploiting the capabilities of each component. The dynamic features of the transient RF signals are captured by GLCT, which offers a reliable and insightful representation of the signals. CNN uses GLCT-processed data to extract intricate spatial characteristics. To accurately represent the developing RF fingerprints, bidirectional-GRU captures the temporal dependencies within the derived features. By combining these strengths, we hope to dramatically increase the accuracy and resilience of RF device identification, opening new opportunities for secure communication, network management, and device authentication.

## 3 Materials and Methods

### 3.1 Materials

To perform wireless device identification, the collection of representative datasets for predictive modeling is an important task. The task involves setting up wireless devices to emit signals that are received and recorded at another device. The signals recorded at the receiving devices are processed to perform feature extraction and data preparation for predictive modeling. The data used in this study is presented by [19] and the process used for its collection, feature extraction, and preparation for predictive modeling is described further. The dataset is generated using nine Nordic IoT devices that are similar to each other as the objective is to perform classification in an intra-model scenario. All the devices broadcast on the 2.4GHz ISM (industrial, scientific, and medical) bands and they are all connected to the same MySensors network. To collect the signal from these devices, a Software Defined Radio (SDR) N200 receiver belonging to the Universal Software Radio Peripheral (USRP) family has been configured with a reference signal with a 10MHz sample rate. Reproducibility in the data collection process is made possible because the XCVR2450 front end of the SDR receiver is locked to the GNSS receiver and regulated to a reference clock of 10MHz. The process of signal acquisition and feature extraction are depicted in Figure 1.

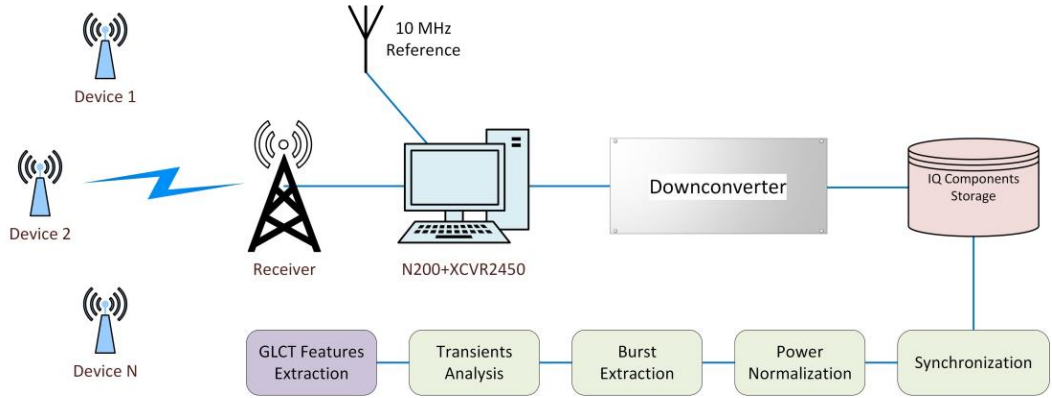

**Figure 1: Dataset Construction for wireless device identification [19]**

3.1.1 Feature Extraction

The process of wireless device identification involves several stages of feature extraction and data preparation. Initially, the signals emitted by nine IoT wireless devices are collected using an SDR. The collected signals, in real-valued format, are downsampled to the baseband and stored as In-phase and Quadrature (IQ) components. To extract the bursts of traffic associated with each device's payload, the signals are synchronized and normalized using power normalization, which involves factoring the signal bursts with their total Root Mean Square (RMS) level.

From each wireless device, 800 bursts are selected, resulting in a total of $9 \times 800 = 7200$ bursts. The transient components of these bursts are then extracted using a moving variance technique and synchronized through cross-correlation. Subsequently, the GLCT [10] is applied to the transients to perform feature extraction, capturing their time-frequency characteristics. Figure 2 provides a depiction of a power signal of a typical wireless device showing the transient portion of the waveform that is used for feature extraction.

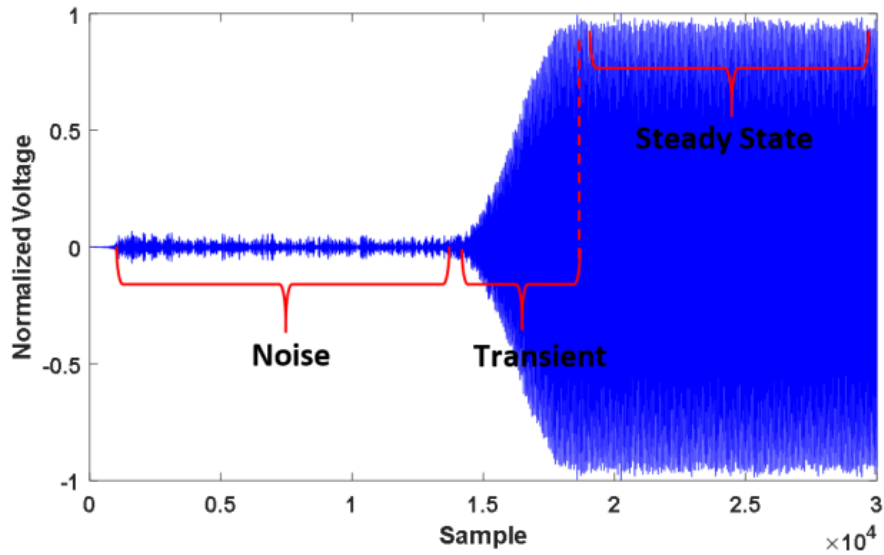

**Figure 2: Depiction of power-on signal of a wireless device [26]**

3.1.2 General Linear Chirplet Transform

The GLCT [10] is a TFA method that extends the Linear Chirplet Transform (LCT) [10]. GLCT

effectively represents multi-component signals with distinct non-linear features, making it suitable for Specific Emitter Identification (SEI) based on transients. It exhibits low sensitivity to noise and belongs to the family of parameterized TFA methods.

GLCT modifies the standard STFT by introducing a demodulated operator to eliminate the influence of the modulated element. This is achieved by incorporating a parameterized rotation in the time-frequency plane, denoted by α. The modified GLCT representation of the signal in the TF domain is obtained by dividing the TF plane into $N + 1$ sections based on the values of α. Mathematically, the GLCT operation can be expressed as Eq. 1 [10]:

$$GLCT(t',w,\alpha) = \int_{-\infty}^{\infty} w(u-t')s(u)e^{-iwu}e^{-i\frac{\tan(\alpha)f_s}{2T_s}(u-t')^2} du \qquad (1)$$

Where:
- $s(t)$ is the input signal.
- $w(u - t')$ represents the window function centered at t'.
- $s(u)$ denotes the input signal s(t) under the window.
- $e^{-iwu}$ is the complex exponential for frequency modulation.
- $\omega$ is the angular frequency.
- $\alpha$ controls the orientation or tilt of the chirplet.
- $f_s$ is the sampling frequency.
- $T_s$ is the sampling period.

The parameter $\alpha$ is defined as Eq. 2 [10]:

$$\alpha = \arctan(\frac{2.T_s}{F_s}c) \qquad (2)$$

Here, $c$ represents the chirp rate. By varying $\alpha$, the TF plane is partitioned into $N + 1$ sections, allowing GLCT to capture different TF characteristics of the signal which can be given through (3):

$$\alpha = -\frac{\pi}{2} + \frac{\pi}{N+1}, -\frac{\pi}{2} + 2\frac{\pi}{N+1}, \dots, -\frac{\pi}{2} + N\frac{\pi}{N+1} \qquad (3)$$

When applying GLCT to SEI, two hyperparameters need to be determined empirically: the window size ($w$) and the number of chirplets ($N$). Optimization of $w$ is important to capture TF characteristics for SEI but can be expensive to compute. An efficient alternative method is used for the optimization of $w$ based on the analysis of signal characteristics in the time domain. By analyzing the standard deviation of transient shapes, a value of $w$ is identified that maximizes the standard deviation across windows of the same size. This optimization method [34] considers the maximum standard deviation for each window size separately. In addition, the method is improved by applying a smoothing filter to the data before computing the standard deviations.

3.1.3 Hyperparameter Optimization for Feature Extraction

The performance of the GLCT is highly dependent on two fundamental hyperparameters: the window size ($w$) and the number of chirplets ($N$). These parameters play a pivotal role in determining the effectiveness of GLCT in capturing the time-frequency characteristics essential for identifying specific emitters accurately.

The window size ($w$) determines the duration of the signal segments analyzed by GLCT, while the number of chirplets ($N$) controls the granularity of the time-frequency representation. However, directly determining the optimal values for $w$ and $N$ can be computationally intensive due to the vast parameter space involved.

To address this challenge, an optimization approach is employed to fine-tune these hyperparameters effectively. One crucial aspect of this optimization process involves evaluating the standard deviation of transient shapes within different windows. This metric provides insights into the variability of signal patterns and is instrumental in determining the optimal window size for analysis. The formula of Eq. 4 is instrumental in calculating the standard deviation of transient shapes within the specified windows. It quantifies the spread of transient shapes, providing valuable information about the diversity and complexity of signal patterns within each window.

$$SD(w) = \sqrt{\frac{\sum_{i=1}^{N}(x_i - \bar{x})^2}{N}} \qquad (4)$$

where:
- $w$ is the window size
- $N$ is the number of chirplet
- $x_i$ is the i$_{th}$ sample in the window
- $\bar{x}$ is the average value of all samples in the window

To develop an effective optimization strategy, a time-domain analysis of signal properties is employed which can allow the process to effectively capture relevant time-frequency characteristics for accurate emitter identification. This process comprises three key steps:

**Computation of Standard Deviation for Each Window Size**: For each window of a given size, the standard deviation of transient shapes is computed. This quantifies the variability of signal patterns within each window.

**Adjustment of Window Size**: The size of each window is adjusted to maximize the standard deviation for each window. A higher standard deviation is preferred as it indicates the presence of diverse and informative transient patterns within that window size, aiding in emitter identification.

**Application of Smoothing Filter**: The data is smoothed using a moving average filter before the standard deviation is calculated. This moving average is chosen for smoothing due to its simplicity and effectiveness in reducing noise while preserving essential signal features, which improves the precision of the optimization process. In contrast to averaging across all windows, which may obscure specific deviations in transient shapes, the moving average gives equal weight to each data point within the window. This ensures that all data points contribute equally to the smoothed value, offering a simple but effective method of noise reduction. Moreover, the study used the moving average filter before computing standard deviations to improve the identification of meaningful transient features in the signal, thereby increasing the overall effectiveness of the optimization approach.

### 3.1.4 Data Preparation

Data preparation is critical, and it is influenced by the learning algorithm used for the classification task [55, 56]. In this case, GLCT features are extracted from IQ components to generate the data. The final dataset is stored in a three-dimensional matrix format, which ensures that there are no duplicates or missing values in the data. To build the model, categorical encoding is applied to the assigned class labels, and the feature set is reshaped into a $1080 \times 900$ matrix. In this case, 900 denotes the features that were extracted for each sample, and 1080 stands for the samples that were collected from nine IoT devices.

To partition the data for model validation, we used 10-fold cross-validation to divide it into ten parts. Nine parts are used for model training and one part is reserved for model validation during each iteration. This procedure makes the most use of the available data for training while ensuring a thorough assessment of the model's performance. This procedure is repeated ten times, using 108 samples for validation and 972 samples for training during each iteration. The final performance evaluation is then calculated by averaging the performance attained during each iteration. The cross-validation procedure is shown in Figure 3, where the performance is averaged over ten iterations and one partition is used for model validation each iteration.

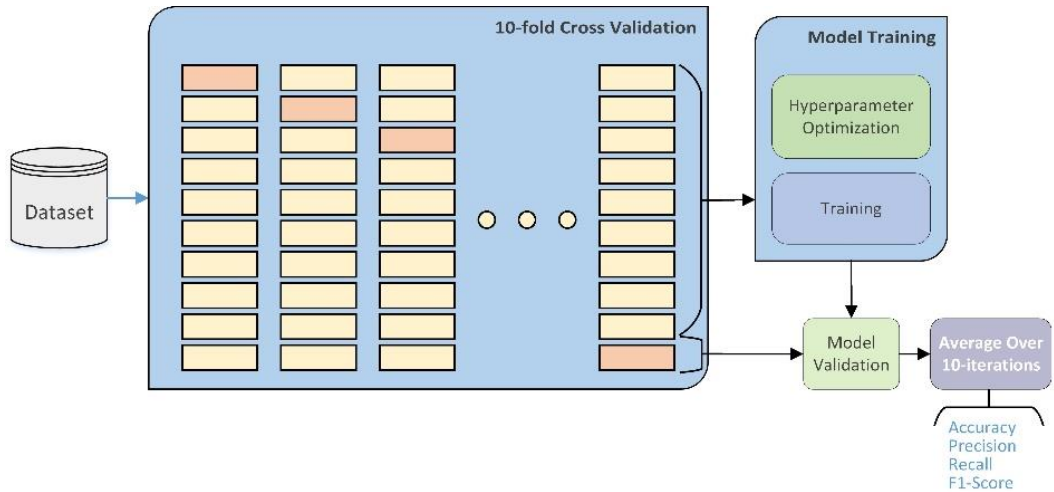

Figure 3: Depiction of the Model building process with 10-fold cross-validation

*3.2 Model Building*

The development of an effective solution for wireless device identification primarily revolves around the model-building stage, which is optimized through the preceding and subsequent stages. Selecting a suitable algorithm for modeling is not governed by a strict rule but relies on intuition, extensive literature study, and experimentation with various algorithms. As our problem involves classification, we have relied on deep learning-based classification algorithms due to their suitability in modeling such problems. Prior research has demonstrated that deep learning-based solutions yield superior performance in RF fingerprinting for wireless device identification [16-18]. Additionally, since the training samples are relatively smaller, it is crucial to construct deep models with a reduced number of neurons to prevent overfitting and enhance generalization performance.

In this study, we conduct experiments with four variants of classification models: (i) CNN, (ii) LSTM, (iii) GRU, and (iv) Hybrid Architecture.

The modeling process involves utilizing the first three deep learning-based classifiers, while the fourth model is constructed by integrating CNN with LSTM and CNN with GRU. As the device identification is based on offline data, the implementation incorporates the bi-directional configuration of LSTM and GRU. These variations in model architecture aim to harness the strengths of different neural network architectures and improve the accuracy of wireless device identification.

3.2.1 Convolutional Neural Networks

The architecture of the optimized CNN architecture used to model wireless device identification is depicted in Figure 4. As the dataset size is small and therefore shallow architecture with dropout is designed to overcome the problem of overfitting. Cross-validation has allowed the identification of the most appropriate architecture for modeling. The model is compiled using Adam optimizer [57] due to its adaptive weight optimization property and suitability for this problem. Categorical cross entropy is used as a loss function as we have used category index instead of one hot encoding for device labelling which is a resource-efficient implementation. An initial learning rate of $3 \times 10^{-3}$ is used and training is carried out for 30 epochs with early stopping based on validation patience with a minimum delta of 0.0001.

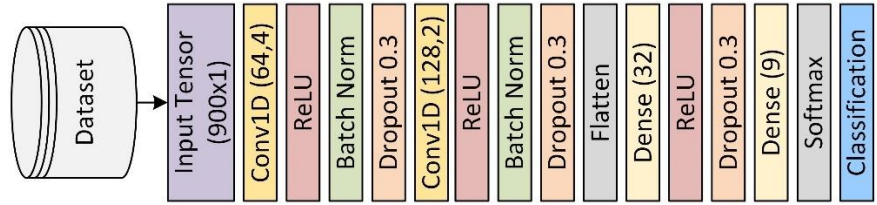

**Figure 4: CNN architecture used for modeling wireless device identification.**

### 3.2.2 Long Short-Term Memory

The architecture of the LSTM model constructed for wireless device identification is depicted in Figure 5. The number of layers and LSTM units in each layer as well as the dropout ratio and number of neurons in dense layers are iteratively identified through cross-validation. The use of bidirectional LSTM units is motivated by the fact that the dataset is relatively small and using a bidirectional LSTM can potentially enhance the model's ability to extract meaningful features by leveraging information from both directions. The model is compiled using RMSProp as an optimizer with categorical cross entropy as a loss function. The initial learning is set to $1 \times 10^{-3}$ and is trained for maximum epochs of 30. Early stopping on validation loss with a minimum delta of 0.01 is used to avoid overfitting.

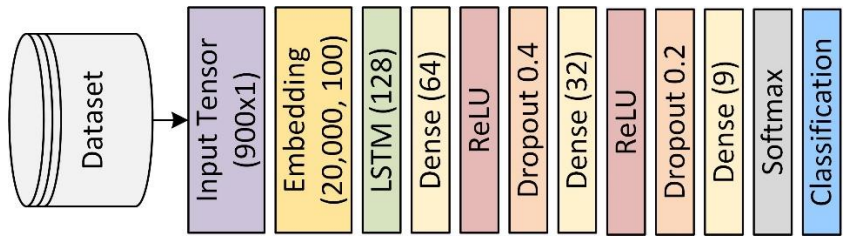

**Figure 5: LSTM architecture used for modeling wireless device identification.**

### 3.2.3 Gated Recurrent Unit

The architecture of the GRU model constructed for wireless device identification is depicted in Figure 6 and is very similar to LSTM. The number of layers and LSTM units in each layer as well as the dropout ratio and number of neurons in dense layers are iteratively identified and the use of bidirectional GRU units is motivated by the fact that the dataset is relatively small and using a bidirectional GRU can potentially enhance the model's ability to extract meaningful features by leveraging information from both directions. The model is compiled using RMSProp as an optimizer with categorical cross entropy as a loss function. The initial learning is set to $1 \times 10^{-3}$ and is trained for maximum epochs of 30. Early stopping on validation loss with a minimum delta of 0.01 is used to avoid overfitting.

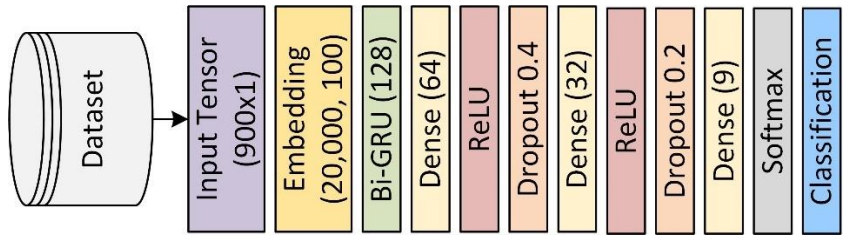

**Figure 6: GRU architecture used for modeling wireless device identification.**

### 3.2.4 Hybrid Architecture

The hybrid architecture is constructed by forming a model with both one-dimensional CNN and memory networks (LSTM and GRU) [58-60]. The final configurations with the best validation accuracy are provided in Figure 7. The model is a combination of one-dimensional CNN followed by two bidirectional GRU layers to perform classification modeling. The designed model incorporates both dropout layers and recurrent dropout layers to reduce the risk of overfitting and improve generalization.

The dropout layer randomly drops connections by setting them zero during training steps whereas the recurrent dropout applies to recurrent layers of the GRU. Like other architectural units, the dropout rates are empirically selected by iterating a set of dropout rates. The model is compiled with an Adam optimizer with an initial learning rate of $1 \times 10^{-3}$ and categorical cross entropy as a loss function. The training is carried out for a maximum of 30 epochs with early stopping based on validation patience with a minimum delta of 0.0001.

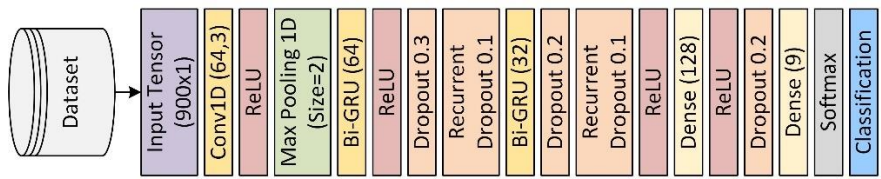

**Figure 7: Hybrid (CNN-bidirectional-GRU) architecture for wireless device identification**

3.2.5 Network Complexity

To provide a comprehensive understanding of the neural network architectures employed in this study, a summary table outlining the complexity of each model in terms of network depth, model size and number of trainable parameters is provide in Table 1.

**Table 1** Network complexity of the four candidate models

| Network | Depth | Size | Parameters |
| --- | --- | --- | --- |
| CNN | 11 layers | 61.06 MB | 15,689 |
| LSTM | 8 layers | 0.765 MB | 199,449 |
| GRU | 8 layers | 0.796 MB | 207,129 |
| CNN-Bi-GRU | 14 layers | 0.321 MB | 83,529 |

The depth of each architecture reflects the number of layers, indicating the level of abstraction and feature representation. Model size, measured in megabytes, indicates storage requirements. Variations in depth and parameter count affect computational complexity and resource usage. The CNN-Bi-GRU has the highest depth at 14 layers, yet it has a smaller size compared to others. This highlights the importance of considering both depth and parameter count. Despite fewer layers, LSTM and GRU have higher parameter counts, suggesting greater internal complexity. This analysis suggests that although the CNN model employed in this study has the smallest model size in terms of the number of trainable parameters, it actually requires higher storage capacity. On the other hand, the CNN-Bi-GRU has an optimized number of parameters and model size, benefiting from increased network depth, which enhances its level of abstraction and feature representation, leading to superior classification performance.

## 4 Results and Discussion

### 4.1 Evaluation Metrics

The purpose of an evaluation of a model's efficacy is to provide a quantitative measure of its performance. The model validation is performed using validation data, and the extent of deviation from the ground truth is measured using several performance measures. It is a classification challenge to determine the identification of a wireless device from its RF signature. Therefore, the model's predictions on the validation data must be utilized to determine the accuracy, precision, recall, and f1-score. Classification accuracy is used as the final criterion for model selection since it is appropriate for the balanced dataset size and relative value of accurate prediction. Precision, recall, f1-score, and accuracy are commonly used metrics for classification evaluation [56].

*4.2 Execution Environment*

The execution environment plays a crucial role in the development and evaluation of the proposed architecture. A combination of carefully selected libraries and tools provides the necessary resources and functionalities for efficient model development and analysis. Moreover, the use of specified versions of libraries and resources used in the modeling process will facilitate the replication of the stated experimental results. Table 1 presents the detailed specifications of the key libraries with their versions that are used for this implementation. These libraries form a comprehensive execution environment that empowers the implementation and evaluation of the proposed architecture.

**Table 2 Execution environment and libraries used in model building and evaluation process.**

| Library | Purpose | Version |
|---|---|---|
| Python 3.7 | Programming language | 3.7 |
| Google Colab | Preferred Python IDE with GPU resources | - |
| Numpy | Support for arrays, Fourier transform, and more | 1.19.3 |
| Pandas | Working with numerical tables and time series | 1.4.4 |
| Matplotlib | Charting library for static and interactive plots | 3.5.3 |
| Seaborn | Enhanced data visualization | 0.11.2 |
| Scikit-learn | Machine learning toolkit | 0.21 |
| Keras | Framework for convolutional neural networks | 3.7.2 |

*4.3 Classification Performance*

The quantitative evaluation results for the four deep learning-based architectures are presented in Table 2 in the form of percentage with precision up to two decimal points. The performance is assessed through cross-validation, and the average performance over 10 iterations is reported for comparison. To assess the performance at various SNR levels, the experiments are conducted at an SNR of 10dB, 20dB, and 30dB. It is noteworthy that the classification accuracy of wireless device identification becomes very high when the SNR of the received signal is kept at 30dB. In the case of CNN and CNN-Bi-GRU models, the classification accuracy reaches 100%. However, the effect of noise on the CNN-Bi-GRU model is not significant and it provides a high classification accuracy of 99.17% even when the SNR is 10dB.

Figure 8 provides a visual representation of the classification performance in the form of a bar chart, indicating that all classifiers demonstrate similar effectiveness in addressing the problem. Particularly, the CNN-Bi-GRU architecture, which incorporates bidirectional units and learns in both space and time, showcases the ability to capture task-specific features with limited training data. By employing regularization techniques such as dropout and early stopping, the models are made robust against various interferences, including background noise. The performance metrics, including precision, recall, F1-score, and accuracy, are presented for each model in the table.

The bar charts for accuracy, precision, recall, and f1-scores for each of the four models for comparison indicate that all the classification models have demonstrated high predictive performance at various SNR levels. The CNN-Bi-GRU architecture, in particular, has the highest predictive performance which can be attributed to their bidirectional units and learning in both space and time. It showcases its ability to capture task-specific features with limited training data. By employing regularization techniques such as dropout and early stopping, the models are made robust against various interferences [61], including background noise indicated by the classification performance at an SNR of 10dB.

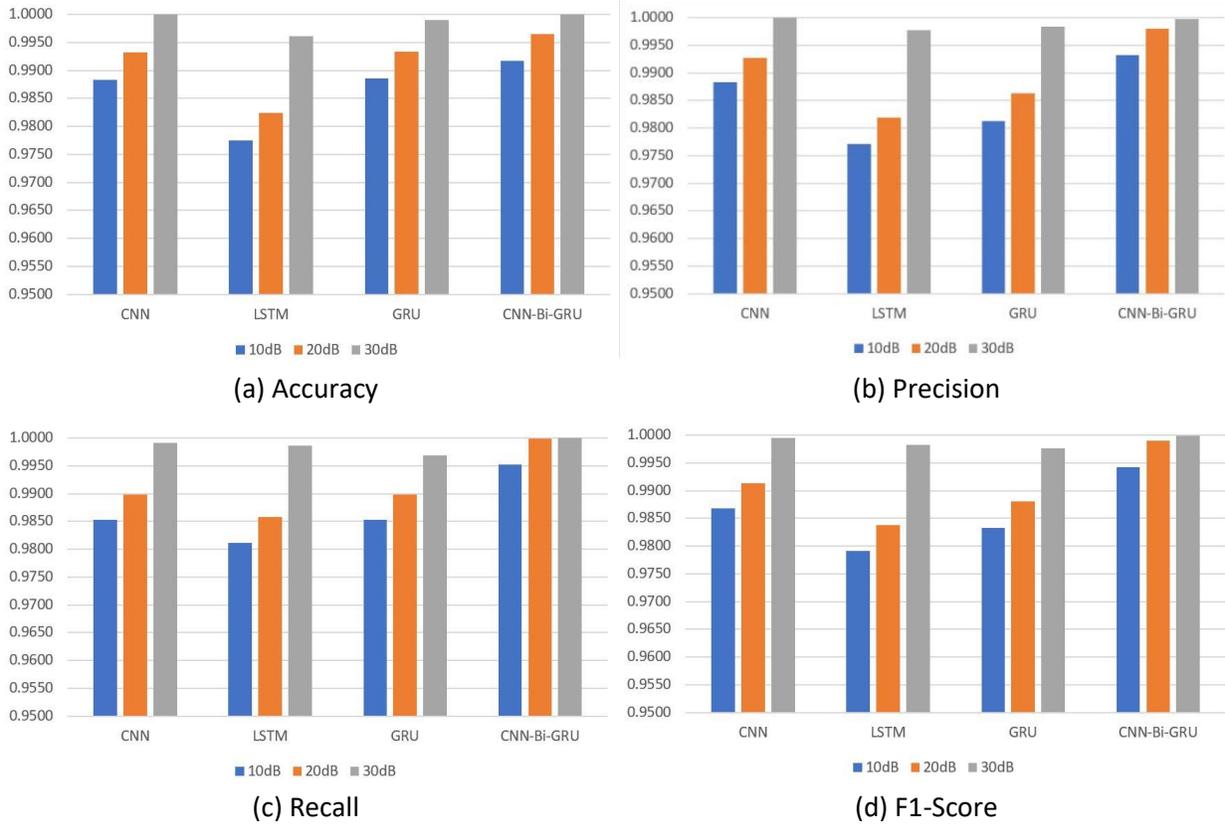

(a) Accuracy  (b) Precision  (c) Recall  (d) F1-Score

**Figure 8 Comparison of models at different SNR levels**

**Table 3 Classification performance of the candidate models**

| Models | Accuracy | Precision | Recall | F1-Score |
|---|---|---|---|---|
| SNR 10dB | | | | |
| CNN | 98.84% | 98.83% | 98.53% | 98.68% |
| LSTM | 97.75% | 97.72% | 98.12% | 97.92% |
| GRU | 98.85% | 98.13% | 98.53% | 98.33% |
| CNN-Bi-GRU | 99.17% | 99.33% | 99.53% | 99.43% |
| SNR 20dB | | | | |
| CNN | 99.32% | 99.28% | 98.98% | 99.13% |
| LSTM | 98.25% | 98.19% | 98.58% | 98.38% |
| GRU | 99.34% | 98.63% | 98.98% | 98.80% |
| CNN-Bi-GRU | 99.65% | 99.80% | 99.99% | 99.90% |
| SNR 30dB | | | | |
| CNN | 100.00% | 100.00% | 99.91% | 99.95% |
| LSTM | 99.61% | 99.77% | 99.87% | 99.82% |
| GRU | 99.90% | 99.84% | 99.68% | 99.76% |
| CNN-Bi-GRU | 100.00% | 99.98% | 100.00% | 99.99% |

*4.4 Comparison with Existing Work*

Regarding RF device identification at different SNR levels, the CNN-Bi-GRU model with GLCT features performs better than existing approaches. Table 3 shows that the suggested method achieves 100% accuracy at 30dB, which is the highest accuracy as compared to lower SNR levels. This result demonstrates how well GLCT's robust feature extraction can be combined with the CNN-Bi-GRU architecture's temporal learning capabilities.

**Traditional methods:** ATT [34] and TE [34] exhibit moderate accuracy at high SNRs but struggle at lower levels (55.70% and 56.40% at 10dB).

**Deep learning approaches:** MD-CNN [62] performs well at high SNRs (86% at 10dB and 92% at 20dB) but falls short of our model's accuracy.

**TFA-based methods:** While OOCW [63], SWVD [64], WVD [64], and CWT-G [64] achieve similar or slightly lower accuracy at 30dB (95% - 99.18%), our model outperforms them at all other SNRs, demonstrating its superior generalizability.

**Table 4 Comparison of existing methods based on classification accuracy.**

| Method | SNR=10dB | SNR=20dB | SNR=30dB |
|---|---|---|---|
| ATT [34] | - | 55.70% | 94.60% |
| TE [34] | - | 56.40% | 77.30% |
| MD-CNN [62] | 86% | 92% | - |
| OOCW [63] | 95% | 98% | 99% |
| SWVD [64] | 95.95% | 98.13% | - |
| WVD [64] | 97.15% | - | 99.18% |
| CWT-G [64] | 97.42% | - | 98.67% |
| CNN-Bi-GRU | 99.17% | 99.65% | 100.00% |

This comparison demonstrates the strengths of our proposed approach, especially its robustness to noise and capacity to capture the complex temporal dynamics of radio frequency signals. Its accuracy and dependability in identifying RF devices in diverse real-world situations renders it, in our opinion, a useful tool.

*4.5 Discussion*

The evaluation of the candidate models reveals important insights into their predictive performance. Among the deep learning-based architectures considered, the hybrid model consisting of a Convolutional Neural Network (CNN) and Bidirectional Gated Recurrent Units (Bi-GRU) stands out as the most superior option. This hybrid architecture offers a unique advantage by incorporating both spatial and temporal learning. The CNN component enables effective feature extraction from the RF signatures, capturing spatial patterns and highlighting important device-specific characteristics. On the other hand, the Bi-GRU component enhances the model's ability to learn temporal dependencies, enabling the capture of sequential patterns in the data. This combination empowers the model to leverage both spatial and temporal information, resulting in more comprehensive and accurate device identification.

The success of the hybrid model is further enhanced by the careful use of dropout, early stopping, and hyperparameter search. Dropout prevents the network from relying too heavily on specific neurons therefore reducing overfitting and improving the generalization capabilities of the model. Early stopping, another regularization strategy, halts the training process when the model's performance on the validation set starts to deteriorate, thus preventing overfitting and ensuring the model is not overly specialized to the training data. Hyperparameter search on the other hand performs the selection of optimal training and architectural parameters from a set of parameters iteratively. These modeling strategies play a crucial role in optimizing the model and mitigating the risk of overfitting, particularly when dealing with a relatively small dataset size.

The significance of these regularization techniques becomes particularly evident during cross-validation. Cross-validation involves evaluating the model's performance across multiple iterations, utilizing different subsets of the data for training and validation. The hybrid model showcases improved performance during cross-validation as compared to other alternatives. The modeling strategies allow the model to maintain its accuracy and reliability across different iterations, ensuring its robustness and generalizability. The ability to consistently perform well in cross-validation demonstrates the model's effectiveness in identifying wireless devices and strengthens its potential for real-world deployment.

In terms of SNR of received signal, classification experiments are performed at three SNR levels of 10dB, 20dB, and 30dB. At a high SNR of 30dB, the classification performance of all the models becomes very high with CNN and CNN-Bi-GRU model providing 100% classification accuracy. In the case of a low SNR of 10 dB, the CNN-Bi-GRU model can accurately identify the wireless devices with an accuracy of 99.17 % and an f1-score of 99.43%.

### *4.6 Challenges and Opportunities for Future Work*

**Dataset Size:** The study employs a meticulously curated dataset; however, the model's applicability to novel scenarios or devices may be limited by the sheer volume and diversity of the data. The robustness and applicability of the model could be further enhanced in subsequent work by gathering and incorporating a larger and more diverse dataset.

**SNR Range:** The study focuses on three specific SNR levels (10dB, 20dB, 30dB). Although the model works well across the board, it may not perform as well in situations with noticeably higher or lower SNR. Given that the results show declining accuracy, future research could focus on extending the model's applicability to a lower SNR range.

**Feature Extraction Techniques:** The GLCT method is used in the study to extract features. Even though they work well, additional methods or combinations of methods could be worthwhile to investigate in order to enhance performance or extract different types of features from the RF signals.

**Real-world Deployment Considerations:** Although the study shows the model's potential for practical applications, actual deployment would probably necessitate other factors like integrating the model into existing systems, resolving latency issues, and making sure it is robust to environmental factors.

### 5 Conclusion

This study focuses on the development of an efficient and robust solution for wireless device identification using RF fingerprinting through the GLCT. The GLCT technique is utilized to extract discriminative features from the transient signal characteristics of RF devices. These extracted features serve as the basis for classification modeling, where four deep learning-based classification models are trained on a carefully prepared dataset. The objective is to predict the device responsible for transmitting an RF signal, enabling device authentication, and ensuring security at the physical layer. The constructed classification models encompass a CNN, bidirectional LSTM, bidirectional GRU, and a hybrid model combining CNN and bidirectional GRU (CNN-Bi-GRU). Remarkably, all candidate models exhibit high prediction performance, demonstrating their effectiveness in wireless device identification. However, among the alternatives, the CNN-Bi-GRU model stands out as superior due to its robust performance and efficient spatiotemporal learning capabilities. The final model achieved exceptional results, with a precision of 99.33%, recall of 99.53%, F1-score of 99.43%, and classification accuracy of 99.17%. These metrics emphasize the model's high accuracy and reliability in predicting the identity of wireless devices. Such impressive performance ensures the potential deployment of the model for real-world applications requiring device authentication and secure communication.

## Declarations

**Ethical Approval**: Not applicable.

**Competing interests*:* The authors declare that there is no conflict of interest.

**Authors' contributions:** The contributions of the authors are listed below:
N. Ahmed: Conception, coding, experimentations and writeup
G. Saleem: Coding, experimentation, writeup and review.
H. M. S. Asif: Conception, supervision, proofread and review.
M. U. Younus: Coding, experimentations, writeup and proofread.
K. Safdar: Writeup and proofread.

**Funding**: Not applicable.

**Availability of data and materials**: https://ieee-dataport.org/documents/iot-transient-radio-frequency-signals